\documentclass[sigplan,screen,nonacm]{acmart}

\setcopyright{none}
\usepackage{subfigure}
\usepackage{multirow}
\usepackage[english]{babel}
\usepackage{amsthm}
\usepackage{algorithm2e}
\usepackage{listings}
\usepackage{xcolor}

\definecolor{codegreen}{rgb}{0,0.6,0}
\definecolor{codegray}{rgb}{0.5,0.5,0.5}
\definecolor{codepurple}{rgb}{0.58,0,0.82}
\definecolor{backcolour}{rgb}{0.95,0.95,0.92}

\lstdefinestyle{mystyle}{
  commentstyle=\color{codegreen},
  keywordstyle=\color{magenta},
  stringstyle=\color{codepurple},
  basicstyle=\ttfamily\footnotesize,
  breakatwhitespace=false,         
  breaklines=true,                 
  captionpos=b,                    
  keepspaces=true,                                   
  numbersep=5pt,                  
  showspaces=false,                
  showstringspaces=false,
  showtabs=false,                  
  tabsize=2
}

\begin{document}
\settopmatter{printacmref=false}

\title{Colossal-Auto: Unified Automation of Parallelization and Activation Checkpoint for Large-scale Models }

\author{Yuliang Liu$^{a}$, Shenggui Li$^{a}$, Jiarui Fang$^{a}$, Yanjun Shao$^{a, b}$, Boyuan Yao$^{a, b}$, Yang You$^{*, c}$}
\affiliation{
 \institution{HPC-AI Technology Inc.$^a$ \hspace{0.3em} Fudan University$^b$ \hspace{0.3em} National University of Singapore$^c$}
}

\thanks{The work was done when Mr. Shao and Mr. Yao was an intern of HPC-AI Technology Inc.}
\thanks{* Corresponding Author (youy@comp.nus.edu.sg). Yang You is a faculty member at NUS; this work was done at HPC-AI Technology Inc.}

\begin{abstract}

In recent years, large-scale models have demonstrated state-of-the-art performance across various domains. 
However, training such models requires the use of various techniques to address the problem of limited computing power and memory on devices such as GPU. 
Some commonly used techniques include pipeline parallelism, tensor parallelism and activation checkpointing. 
While existing works have focused on finding efficient distributed execution plans\cite{zheng2022alpa} and activation checkpoint scheduling\cite{herrmann2019optimal}\cite{beaumont2021efficient}, there has been no method proposed to jointly optimize these two plans. 
Moreover, ahead-of-time compilation relies heavily on accurate memory and computing overhead estimation, which is often time-consuming and misleading. Existing training systems and machine learning pipelines either physically execute each operand or estimate memory usage with a scaled input tensor. 
To address these challenges, we introduce a system that can jointly optimize distributed execution and gradient checkpointing plans. 
Additionally, we provide an easy-to-use symbolic profiler that generates memory and computing statistics for any PyTorch model with the minimal time cost. Our approach allows the user to parallelize their model training on the given hardware with minimum code change based. 
The source code is publicly available at \href{https://github.com/hpcaitech/ColossalAI}{ColossalAI}.


\end{abstract}

\maketitle

\section{Introduction}

Foundation models~\cite{bommasani2021opportunities} have emerged as a leading paradigm in the field of AI, with neural networks such as BERT~\cite{bert}, GPT-3~\cite{gpt-3}, and Vision Transformer~\cite{https://doi.org/10.48550/arxiv.2010.11929} containing a large number of parameters and trained on massive datasets. As the number of parameters grows, these models show improved performance, and with the advent of powerful hardware, there has been a trend toward building increasingly larger models. However, this trend has resulted in a situation where a single GPU is unable to meet the high demand for fast memory. Without any optimization, training a model of 10 billion parameters can cost more than 80 GB of memory, even with just one sample per batch. As a result, large models in modern times are trained in a distributed manner, with systems such as GShard~\cite{gshard}, FairScale~\cite{FairScale2021}, Megatron-LM~\cite{shoeybi2019megatron}, and DeepSpeed~\cite{deepspeed} providing strategies for distributed training on limited devices.

However, distributed training of large models requires significant engineering effort to find the optimal execution plan with Single Program Multiple Data (SPMD) style intra-op parallelism~\cite{xu2021gspmd}. This is due to the large search space resulting from the combination of two factors: each dimension of the operator tensor can be partitioned, and devices can be viewed as an arbitrary N-dimensional mesh. Furthermore, the output of one operator is the input of another, which may require a completely different sharding specification. Since it is almost impossible to implement the transformation between two sharding specifications exhaustively, existing methods require that the number of dimensions of the device mesh and tensor partition is less than or equal to 2. Additionally, the search for the best intra-op parallelism can be interfered with by other optimizations, such as activation recomputation~\cite{act-ckpt} and model data offloading strategies~\cite{ren2021zero, fang2022parallel}, which can reduce memory consumption and lead to a completely different search space for intra-op parallelism. Therefore, developing an automatic parallel system is a significant but challenging task to simplify the process of finding the optimal execution plan.

Current automatic intra-op parallelism solutions, such as Alpa~\cite{zheng2022alpa} and Flexflow~\cite{jia2019beyond}, still have some limitations. For example, activation recomputation has become a must-have recipe for foundation model training, but existing methods do not take it into consideration. Additionally, the generalization for new hardware configurations is insufficient, as existing work can only organize devices into a 2D mesh and cannot work for more complex typologies.

In this paper, we present an automatic intra-op parallel system based on PyTorch~\cite{paszke2019pytorch}, which compiles serial model code into efficient parallel execution code. Besides parallelization, our system can automatically generate the most efficient activation recomputation scheme for an execution plan to meet the device memory budget. To achieve this, our system collects information from both hardware and  computation graphs in a profiling phase, adopts a hierarchical optimization method to find the optimal execution plan, and generates runtime execution code. Specifically, our contributions include:

\begin{itemize}
    \item We jointly optimize the intra-op parallelism and activation recomputation methods ahead of time. 
    \item A heuristic algorithm to convert the tensor layout of different sharding specs is developed to support efficient tensor layout conversion in an arbitrary dimension device mesh.
    \item We implement a profiler that could collect both computing and memory overhead in the fine-grained computation graph using meta-execution.
\end{itemize}
\section{Background}

As the memory on common accelerators such as Nvidia GPU is limited, it is challenging to train a large-scale model on one device as the program can easily go out of memory. Thus, distributed training is a necessity to train models beyond the single-device memory limit. Researchers from industry and academia have proposed a wide array of techniques to speed up model training on a GPU cluster. We will discuss these techniques in the following sections.
\begin{figure*}[ht!]
    \centering
    \includegraphics[width=1\textwidth]{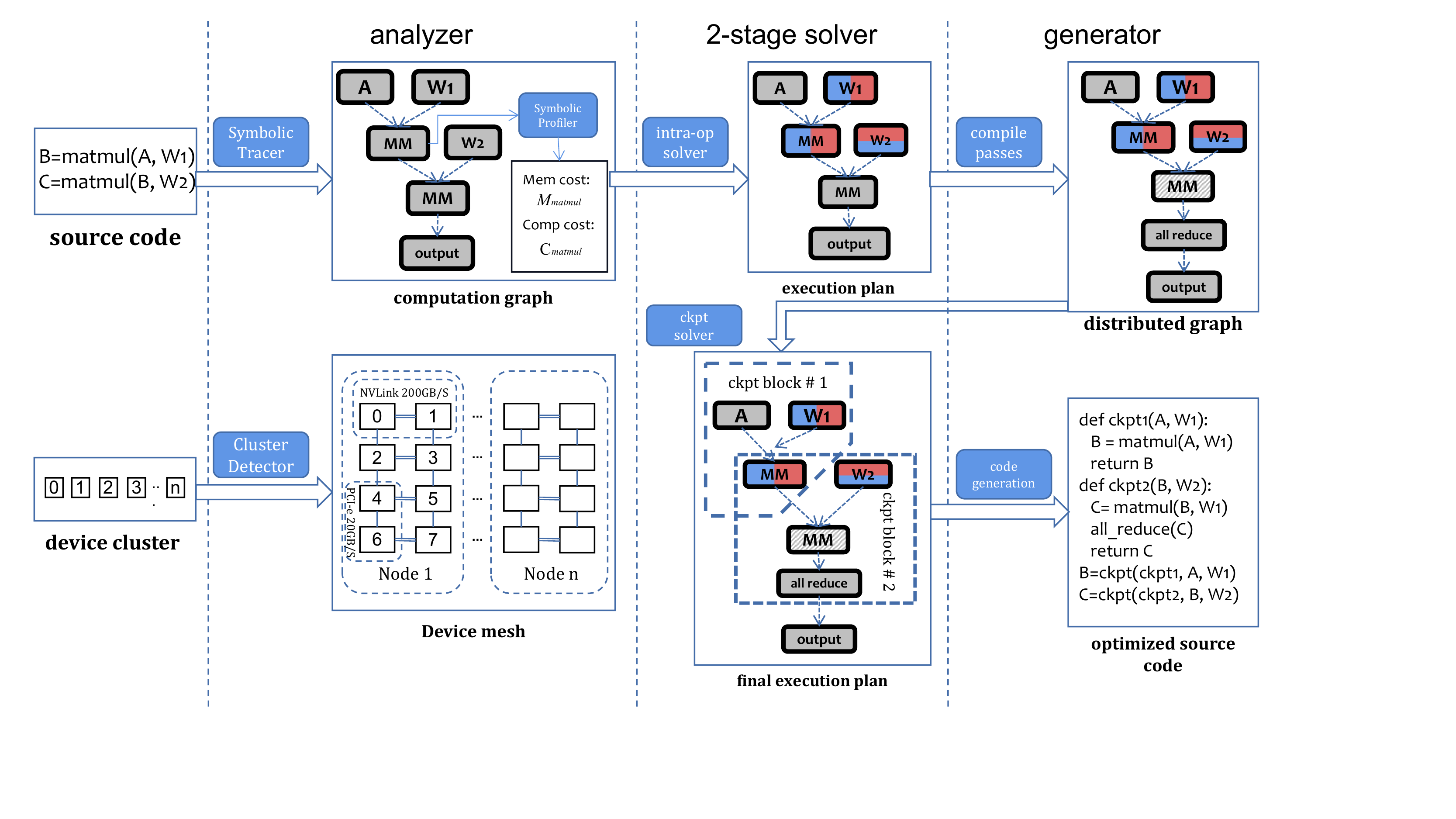}
    \caption{Workflow of Colossal-Auto}
    \label{fig: workflow}
\end{figure*}
\subsection{Data Parallelism}

Data parallel training is the most common way to conduct distributed training due to its simplicity. In data parallel training with $N$ GPUs, the dataset is sharded into $N$ partition and the model is replicated over all $N$ GPUs. The model is trained with the local dataset partition and synchronizes the gradients during the backward pass. Tools such as Horovod~\cite{horovod} and PyTorch DistributedDataParallel~\cite{pytorch-distributed} allow the users to integrate data parallel training into their code with minimum code changes.

One major drawback of data parallelism is memory redundancy since each GPU holds the full copy of the model weights. This restricts the model size scaling as having more GPUs does not allow the user to train larger models. To remove such redundancy, DeepSpeed~\cite{deepspeed} has proposed the Zero Redundancy Optimizer~\cite{rajbhandari2020zero} which partitions the optimizer states, gradients, and model parameters during data parallel training. As a result, the model scale can easily reach billions of parameters in data-parallel training.
Besides, DeepSpeed also offers offloading~\cite{zero-infinity} to move tensors from GPU memory to CPU memory and even NVMe disks when not in use. This further enlarges the amount of available memory to accommodate larger-scale models.

\subsection{Pipeline Parallelism}

Besides sharding the dataset, other works such as GPipe~\cite{gpipe}, PipeDream~\cite{pipedream}, Chimera~\cite{chimera}, and Megatron-LM~\cite{shoeybi2019megatron} proposed to pipeline parallelism shard the model by layer. In pipeline parallelism, the model is split into chunks of consecutive layers and each chunk is allocated to one device. The layer activation and output gradients are passed between pipeline stages during the forward and backward pass respectively. As a result, the model scale can increase with the number of devices, supporting large-scale model training. However, one problem is that there will be bubble time where one device is engaged in computation while others stay idle during training. To tackle this problem, Megatron-LM~\cite{shoeybi2019megatron} proposed interleaved pipelining and Chimera proposed bidirectional pipelining, leading to a significant reduction in the bubble time. 

\subsection{Tensor Parallelism}

Tensor parallelism refers to the technique to shard the model weight and execute training in the SPMD fashion. GShard~\cite{gshard} allows the user to annotate the sharding plan for selected tensors in the computation graph and infer the sharding plan for other tensors using iterative data-flow analysis. Megatron-LM~\cite{shoeybi2019megatron} also proposed tensor parallelism, specifically targeted at the Transformer~\cite{transformer} architecture. Megatron-LM applied row-wise and column-wise sharding to the linear layers in the self-attention module and the feed-forward module. This reduces the memory footprint and the amount of computation on a single GPU, as a result, the model can be scaled up to billions of parameters.

Tensor parallelism is orthogonal to data and pipeline parallelism and they can be used together for distributed training. As the amount of communication is generally smaller in pipeline parallelism and cross-node communication is significantly slower than that of intra-node communication, tensor parallelism is often restricted to intra-node training while pipeline parallelism is used for cross-node training to better utilize the hardware bandwidth~\cite{shoeybi2019megatron}.

\subsection{Automatic Parallelism}

While the techniques mentioned above use static strategies, another way to achieve parallelism is to search for a suitable parallelization strategy for the given model. FlexFlow~\cite{flexflow} and Tofu~\cite{tofu} have proposed different search strategies to parallelize model training. 

However, they are not designed for large-scale model training and deliver sub-optimal performance. The recent advancement comes from Alpa~\cite{zheng2022alpa}, which generates a high-performance parallelization execution plan on a cluster. 
Alpa sees parallelization from a different point of view. It considers both data and tensor parallelism as intra-operator parallelism and pipeline parallelism as inter-operator parallelism. 
It formulates the intra-operator as an ILP problem and searches for the optimal sharding for tensors.
Its inter-operator search is built on TeraPipe~\cite{terapipe} to find the optimal module partitioning using dynamic programming. 





\subsection{Activation Checkpoint}

Activation checkpoint~\cite{act-ckpt, dtr} is a technique to reduce the memory footprint on a single GPU by trading computing for memory. When an activation checkpoint is applied to a group of consecutive layers, only the output of the last layer is cached for the backward computation. All other intermediate outputs are not stored during the forward pass. During the backward pass, re-computation is triggered to obtain the intermediate outputs temporarily for gradient computation. As a result, the memory consumed by intermediate activations can be significantly reduced, making more memory available to accommodate larger models. As this technique does not shard tensors, it stays orthogonal to other parallelization techniques. To decide where to use activation checkpoint in the training pipelines, Rotor~\cite{herrmann2019optimal}, Pofo~\cite{beaumont2021efficient}, Checkmate~\cite{jain2020checkmate}, POET~\cite{patil2022poet} manage to automate this process, finding the best strategies for running time with given memory budget.

\section{Design}

With the wide array of techniques mentioned above, distributed training is made much easier. However, there is no system to integrate all of them to achieve the ultimate automation for parallel training. In the most recent work, Alpa has unified data, tensor, and pipeline parallelism, but not activation checkpoint. Therefore, our work attempts to take activation checkpointing into consideration and generates a near-optimal execution plan which includes both parallelization and activation checkpoint. \textbf{Currently, our work is limited to intra-operator parallelism and inter-operator parallelism will be implemented in the future.}

Colossal-Auto is a system built upon PyTorch FX~\cite{torch-fx}. 
The input of the system is the PyTorch model written in a non-distributed manner, and the output is a transformed PyTorch model that can be executed with efficient intra-op parallelism and automatically selected activation checkpointing for the cluster environment.
As the output model is a PyTorch model, it is compatible with runtime optimization methods, such as ZeRO-Offload~\cite{ren2021zero} and PatrickStar~\cite{fang2022parallel, huang2022elixir}. A sample code snippet is given below.

\begin{lstlisting}[language=Python, label=demo, caption=an example to show how to use our system with \textit{\textbf{one-line changing}} from user source code., label=code sample]
model = resnet50(num_classes=10)

# convert original model to optimized model
model = autoparallelize(model, input_sample)

criterion = torch.nn.CrossEntropyLoss()
optimizer = torch.optim.SGD(model.parameters())

# a normal training loop
for epoch in range(NUM_EPOCHS):
    for img, label in data_loader:
        optimizer.zero_grad()
        output = model(img)
        train_loss = criterion(output, label)
        train_loss.backward(train_loss)
        optimizer.step()
\end{lstlisting}

\section{Analyzer}

The analyzer is a static analysis system built for PyTorch. As PyTorch is a dynamic-graph-based machine learning framework, it is difficult to obtain the graph information before execution. Our analyzer is built upon the PyTorch FX module~\cite{torch-fx} to obtain the static computation graph ahead of time. 

The analyzer consists of three parts: a \textit{symbolic profiler} for collecting computing and memory overhead related to static computation graph, a \textit{cluster detector} for collecting hardware characteristics and detecting cluster topology, and a \textit{tensor layout manager} to find efficient tensor layout conversion the path from different sharding spec and record conversion cost.\par

\subsection{Symbolic Profiler}


In order to evaluate the best strategies for parallelization and activation checkpoint, we need to obtain information such as the number of floating point operations and memory consumption. However, as PyTorch is a machine learning framework based on dynamic graph execution, such information is generally collected during runtime, imposing a significant challenge to our search.

\begin{figure}[h]
    \centering
    \includegraphics[width=0.5\textwidth]{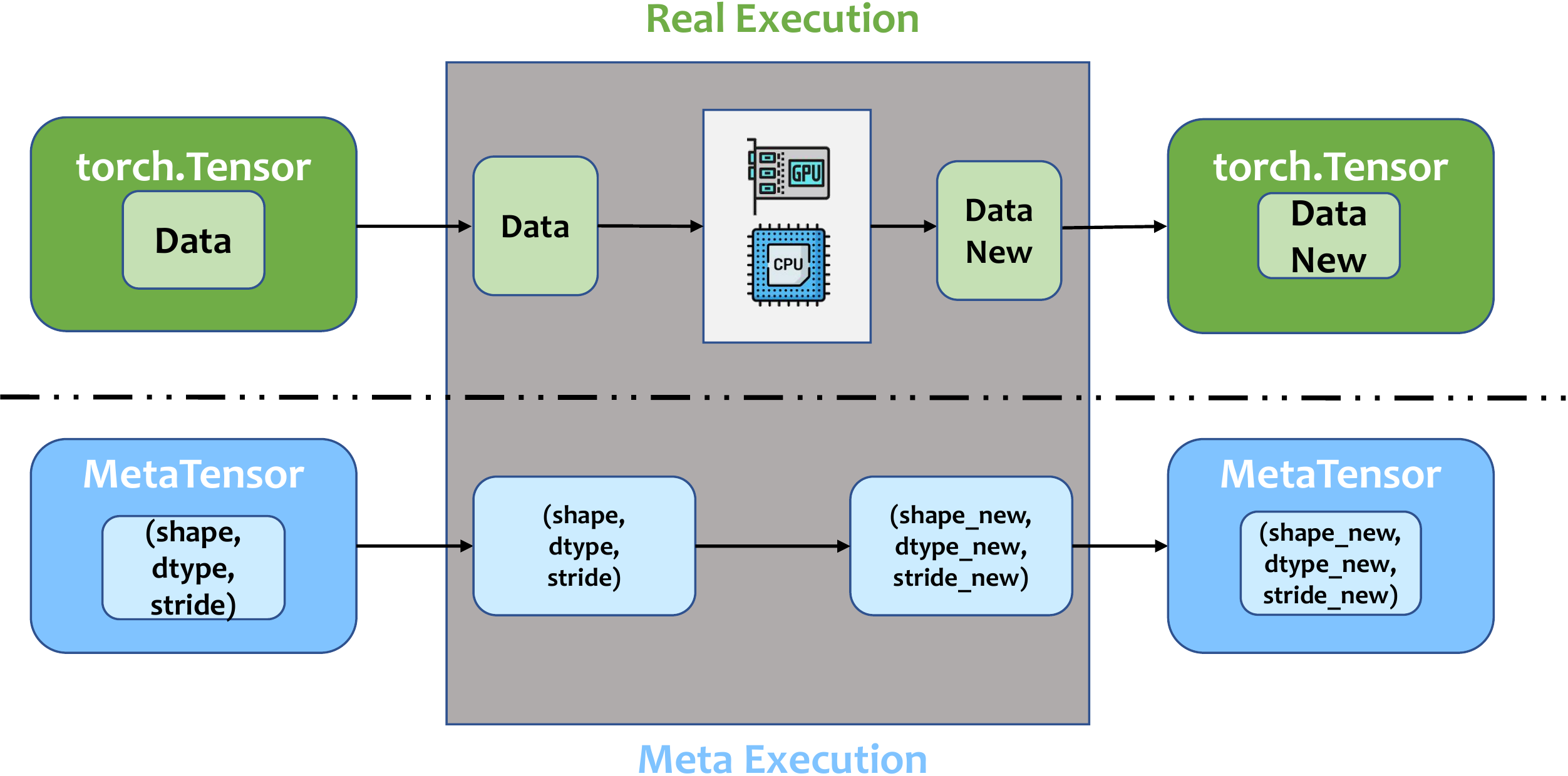}
    \caption{symbolic execution}
    \label{fig: symbolic}
\end{figure}

With the latest PyTorch FX ~\cite{torch-fx} module, this becomes possible with symbolic tracing. Symbolic tracing does not conduct real computation, instead, it only infers the output's meta data such as shape and data type as shown in figure~\ref{fig: symbolic}. We improved upon the PyTorch FX~\cite{torch-fx} symbolic tracing to support a wider range of operations, control flow, and PyTorch meta tensors. During tracing, the analyzer annotates the meta information for each node, and collects both the computing overhead (TFLOPs) and memory overhead (Byte) using symbolic execution.
In this way, since computation is only simulated instead of actually executed, we are able to extract the computation graph and relevant meta information ahead of execution and the overhead of profiling is reasonably low.

\begin{figure*}[h!]
    \centering
    \includegraphics[width=0.9\textwidth]{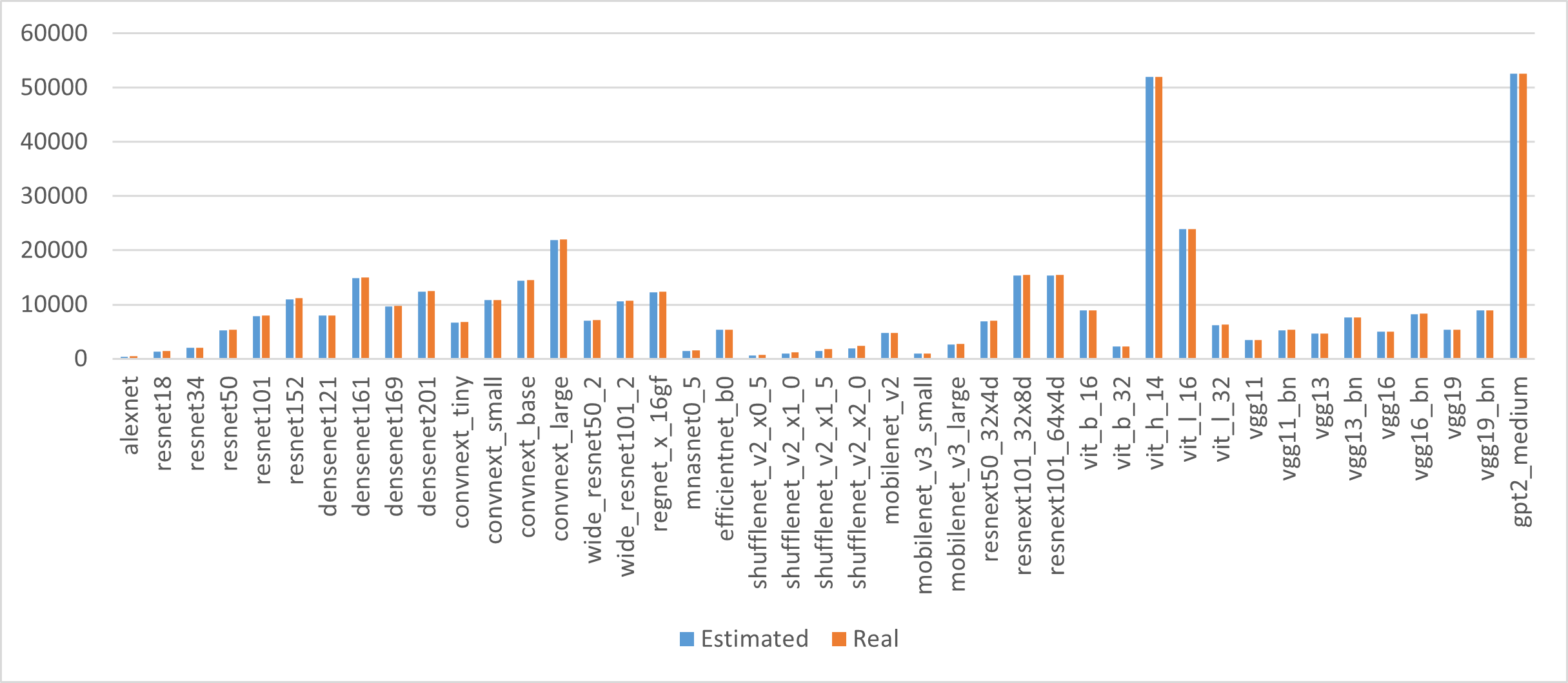}
    \caption{symbolic memory profiling evaluation}
    \label{fig: evaluation}
\end{figure*}

In order to evaluate the accuracy of this profiler, we tested it on more than a dozen mainstream algorithm models including VGG\_16, Resnet50, ViT, GPT2, etc. The memory estimation result is very close to the value of real execution as shown in figure~\ref{fig: evaluation}.

\subsection{Cluster Detector}

We have adopated the concept of device mesh from Alpa~\cite{zheng2022alpa} in our work to have an abstraction for the devices involved in distributed training. The device mesh represents a group of devices as a 2D mesh. For example, 8 physical GPUs can be viewed as 1x8 or 2x4 logical meshes. In the logical device mesh, all computation and collective communication operations will equally occur in each dimension of the device mesh. Therefore, the device mesh need to ensure that devices on the same axis have the same communication performance as each other, otherwise, the device with the lowest bandwidth will slow down the communication process.

Alpa assumes the communication bandwidth between devices in the same computing node are same. However, there are some clusters only have NVLink between 4 pairs of adjacent GPUs. In order to have a better performance, we need to be aware of the fine-grained topology of cluster computing nodes.

To achieve this goal, our cluster detector collects cluster communication performance, such as latency and bandwidth, and detects fine-grained cluster topology. The communication latency is estimated by communicating a series of small-size communication objects in the process group, and the algorithm bandwidth is estimated by communicating large-size communication objects in the process group. With the algorithm bandwidth, the bus bandwidth could be derived through the algorithm bandwidth. With the cluster communication performance, we can deduce the network topology and use it to construct an optimal device mesh where the devices in the same axis have equal communication capability. At the same time, communication performance attributes could be added to the constructed device mesh through the detected cluster information.

\subsection{Tensor Layout Manager}

In intra-op parallelism, a tensor can be sharded into different layouts. Therefore, a representation is needed to describe how a tensor is sharded. We follow Alpa's definition of SMPD-style sharding specifications in our system.
For an N-dimensional Tensor, its sharding spec is to be defined as $X_0X_1...X_{n-1}$, where $X_i$ belongs to {$S$, $R$}, and the representation of this $S$ has a subscript.
If the spec corresponding to $X_i$ is $S_j$, it means that the tensor will be sharded along the $j$th axis of the device mesh in its $i$th dimension. 
In particular, if $S$ has two or more subscripts, it means that the shard will be performed along all subscript axes.

When a tensor is required to have different sharding specs in upstream and downstream operators, we need to perform layout conversion processing, which is called redistribution. There are currently two mainstream methods, enumeration conversion, and dimension-by-dimension conversion. 

Enumeration conversion is to enumerate all possible situations, and then find the corresponding conversion scheme in the table when conversion is required.
However, as the dimension of the device mesh increases, it is difficult to compute the conversion path, and the size of the enumeration is significantly large. 

Dimension-by-dimension conversion is for a convert the sharding spec from its first dimension to the last dimension in the sequence given an N-dimensional tensor. For example, assuming we have a tensor whose sharding spec is $X_0X_1...X_{n-1}$, we need to transform this sharding spec $X'_0X'_1...X'_{n-1}$. We start from the 0th dimension and convert from $X_0$ to $X'_0$ and repeat until the last dimension. In this way, no matter how many dimensions the device mesh and tensor have, a feasible conversion sequence can always be generated. However, the conversion efficiency is poor as it does not guarantee the conversion sequence is optimal.

To tackle these problems, we propose a novel algorithm~\ref{alg: layout conversion}, using heuristic search, to solve the conversion problem of sharding spec, which can be described as:
\begin{enumerate}
    \item Generate all one-step transform sharding specs from source spec
    \item In the one-step transform sharding specs, according to the similarity function, select a sharding spec with the "least difference" as the subsequent source sharding specification, and record the sharding spec in the transform path. If a sharding spec of the one-step transforms is the same as the target sharding spec, the algorithm ends.
    \item Repeat 1, 2 until the end of the algorithm
\end{enumerate}


\RestyleAlgo{ruled}

\begin{algorithm}[h]
\caption{An algorithm with caption}
\label{alg: layout conversion}
\KwData{$source\_spec, target\_spec$}
\KwResult{$transform\_path$}
$total\_step \leftarrow 0$\;
$transform\_path \leftarrow []$\;
$transform\_path[total\_step] \leftarrow source\_spec$\;
\While{$min\_diff \neq 0$}{
$min\_diff \leftarrow \infty$\;
$valid\_transforms \leftarrow Transform(source\_spec) $\;
\ForAll{$sharding\_spec \in valid\_transforms$}{
$diff\leftarrow Heuristic(sharding\_spec, target\_spec)$\;
\If{$diff < min\_diff$}{
$source\_spec \leftarrow sharding\_spec$\;
$transform\_path(total\_step) \leftarrow source\_spec$\;
}
}
$total\_step \mathrel{+}=  1$\;
$transform\_path(total\_step) \leftarrow source\_spec$\;
}
\Return  $transform\_path$\;
\end{algorithm}

\noindent\textbf{One-step transform.} In our layout manager, there are three operations that will change the sharding spec: all gather, shard, and all to all. All feasible sharding specs derived from each operation's single conversions will be added to the one-step transform sharding spec list. For example, the one-step transform list of $S_0R$ is [$RR$, $S_0S_1$, $S_{01R}$, $RS_0$]. Note that there may be some operations that are not valid here because the sizes of different dimensions of the tensor will be different. If the dim spec of Xi is converted to Sj, and the size of the i-th dimension cannot be divided by the j-th axis of the device mesh size, then this sharding spec is invalid.

\noindent\textbf{Heuristic function.} The rule of the heuristic function we used to measure the difference between two sharding specs dimension by dimension, so the difference between source\_spec and target\_spec can be expressed as\\ 
\begin{center}
$\sum_{\substack{i \in (0, 1, \dotsc,dim(s)-1)}} dim\_diff(s[i], t[i])$.
\end{center}

\textit{Dim difference} will calculate the cost of conversion between dimensions accumulate the cost of all operations, and the final value is the total cost. Note that the overhead here does not refer to the actual time overhead or communication volume of the operation, but is a value used to describe the difference between the two dim specs. The operations that can change the dim spec are all-gather and shard. Because shard is an on-chip operation, and all gather communication takes place cross devices, we need to assign a larger cost to the all-gather operation. In addition, if more than one conversion is required between two dim specs, a step penalty will be imposed. For example, $S_0$ -> $S_1$, the conversion path is $S_0$ -> R -> $S_1$, that is, it needs to perform all gather on the 0th axis of the device mesh, and then perform sharding on the 1st axis of the device mesh, so $dim\_diff(S_0, S_1) = cost(all\_gather) + cost(shard) + step\_penalty.$

\noindent\textbf{Solver supports.} In addition to providing the core functions of the above search, our layout manager records the communication overhead required to convert a tensor from source spec to target spec for layout conversion. This overhead is estimated by the alpha-beta cost model, where the alpha and beta values are obtained from the cluster detector and will be used in both the intra-op parallelism solver and activation checkpoint solver.\par

At the same time, in order to avoid repeated searching at runtime, we use the solved source-target spec pair as the key and the corresponding transform path as the value and put it in a \textit{cache dictionary}. Because our execution plan is fixed at runtime, and the intra-op parallelism solver will calculate the tensor layout converting cost for all possible upstream and downstream sharding spec conversions, we will not do any tensor layout converting search during runtime.
\section{Two-Stage Solver}

Our system searches the execution plan through two levels: \textbf{Intra-op parallelism} and \textbf{activation checkpoint}. Such a two-level hierarchical search is beneficial for two reasons. 
Firstly, the strategy search time for intra-op parallelism is positively correlated to the number of nodes in the graph, and the computation graph without activation checkpoint recomputing has fewer nodes than the original computation graph, this greatly reduces our solving time.
Secondly, The previous activation checkpoint work did not consider a distributed model which involves many communication operations, leading to poor performance on a distributed model. 
We have taken the communication overhead into consideration for our activation checkpoint solver so that we can find a better activation checkpoint scheme for the distributed model.

\subsection{Intra-op Parallel Solver}

Our solver is adapted from Alpa's intra-op parallel ILP solver~\cite{zheng2022alpa}, and we implement some engineering tricks to keep generality and reduce the solving complexity of this solver.

As the operations captured by PyTorch FX are not primitive enough, the number of operations is large and makes it complicated to match node to sharding strategies. To reduce the complexity, we created one more abstraction layer of the strategy generator. We constructed a node dispatcher and the node in the computation graph will be dispatched to the corresponding strategy generator according to the operation type. For example, addbmm, bmm, and matmul can use the same strategy generator, and most unary elementwise operators can share the same strategy generator. With this method, we only need to construct no more than 20 strategy generators to support all operators in the GPT2 model.

To further reduce the search complexity, we have implemented the following optimizations. Firstly, thanks to our AoT symbolic profiler and hierarchical searching, we only consider the computation graph of the forward pass, it will significantly reduce graph node size as well as the search complexity. secondly, we merge the computationally-trivial nodes, such as getitem nodes, element-wise nodes, or slice nodes, with the computationally intensive node. Lastly, in the PyTorch FX graph, some nodes have neither tensor object inputs nor tensor object outputs, such as the addition of two scalars. For those nodes, we will remove them from the computation graph. With those optimizations, our solver could solve the problem in a reasonable time. 

\subsection{Activation Checkpoint Solver}

The checkpoint solver will take in the distributed computation graph and search for the best subgraph to apply activation checkpointing.

\subsubsection{Modeling}\label{modeling}

We inherit the \textit{Rotor} algorithm\cite{herrmann2019optimal} for automatic activation checkpointing. This algorithm considers a model as a chain of $L$ stages where a stage could be a single layer or a group of multiple layers, i.e. the linearized neural network. For a particular stage $\ell$, we could compute the forward activation $a^{\ell}$ and backward activation $\delta^{\ell-1} = \frac{\partial \mathcal{L}}{\partial a^{\ell}}$ with the following components

\begin{equation}
\begin{gathered}
F^{\ell}: a^{\ell}=f_{\ell}\left(\theta^{\ell}, a^{\ell-1}\right) \\
B^{\ell}: \delta^{\ell-1}=\bar{f}_{\ell}\left(\theta^{\ell}, \delta^{\ell}, \bar{a}^{\ell}, a^{\ell-1}\right) \\
\frac{\partial \mathcal{L}}{\partial \theta^{\ell}}=\bar{g}_{\ell}\left(\delta^{\ell}, \bar{a}^{\ell}, a^{\ell-1}\right)
\end{gathered}
\end{equation}

where $\theta^{\ell}$ stands for the whole set of parameters involves in the forward computation of stage $\ell$, $\bar{a}^{\ell}$ stands for the set of intermediate activation values that are required to compute the back-propagation inside the block.

After that, we could define multiple operations needed during the activation checkpoint scheduling

\begin{table}[H]
    \resizebox{0.47\textwidth}{!}{
    \centering
        \begin{tabular}{|c|c|c|c|c|}
            \hline
            Operation & Input & Output & Time & Memory overhead \\
            \hline
            \multirow{2}{*}{$F_{all}^{\ell}$} & $\{a^{\ell - 1}\}$ & $\{a^{\ell - 1}, \bar{a}^{\ell}\}$  & \multirow{2}{*}{$u_f^{\ell}$} & \multirow{2}{*}{$o_f^{\ell}$} \\
            & $\{\bar{a}^{\ell - 1}\}$ & $\{\bar{a}^{\ell - 1}, \bar{a}^{\ell}\}$ & & \\
            \hline
            \multirow{2}{*}{$F_{ck}^{\ell}$} & $\{a^{\ell - 1}\}$ & $\{a^{\ell - 1}, a^{\ell}\}$ & \multirow{2}{*}{$u_f^{\ell}$} & \multirow{2}{*}{$o_f^{\ell}$} \\
            & $\{\bar{a}^{\ell - 1}\}$ & $\{\bar{a}^{\ell - 1}, a^{\ell}\}$ & & \\
            \hline
            $F_{\varnothing}^{\ell}$ & $\{a^{\ell - 1}\}$ & $\{a^{\ell}\}$ & $u_f^{\ell}$ & $o_f^{\ell}$ \\
            \hline
            \multirow{2}{*}{$B^{\ell}$} & $\{\delta^{\ell}, \bar{a}^\ell, a^{\ell - 1}\}$ & $\{\delta^{\ell - 1}\}$ & \multirow{2}{*}{$u_b^{\ell}$} & \multirow{2}{*}{$o_b^{\ell}$} \\
            & $\{\delta^{\ell}, \bar{a}^\ell, \bar{a}^{\ell - 1}\}$ & $\{\delta^{\ell - 1}, \bar{a}^{\ell - 1}\}$ & & \\
            \hline
        \end{tabular}
    }
    \caption{List of Operations}
    \label{tab:act_ops}
\end{table}

To combine this activation checkpoint solver with an Intra-op parallel solver, we further consider the communication overhead of each stage, notations are listed below

\begin{table}[H]
    \centering
    \begin{tabular}{|c|c|c|}
        \hline
        Phase & Time & Memory overhead \\
        \hline
        Forward & $u_{fcomm}^{\ell}$ & $o_{fcomm}^{\ell}$ \\
        \hline
        Backward & $u_{bcomm}^{\ell}$ & $o_{bcomm}^{\ell}$ \\
        \hline
    \end{tabular}
    \caption{Communication Overheads}
    \label{tab:comm_ops}
\end{table}

With table \ref{tab:act_ops} and \ref{tab:comm_ops}, we could modify the main theorem in \cite{herrmann2019optimal} into the following one

\begin{theorem}
    $C_{BP} (s, t, m)$, the optimal time for any valid persistent sequence to process the chain from stage $s$ to stage $t$ ($t \geq s$) with available memory $m$. is given by
    \begin{equation}
        \begin{aligned}
            C_{BP}(s, s, m) & = \begin{cases}u_f^s + u_{fcomm}^s + u_b^s + u_{bcomm}^s & m \geq m_{\text {all }}^{s, s} \\
            \infty & m<m_{a l l}^{s, s}\end{cases} \\
            C_{BP}(s, t, m) & =\min \left(C_1(s, t, m), \quad C_2(s, t, m)\right) \\
        \end{aligned}
    \end{equation}
    
    where

    \begin{equation}
        \begin{aligned}
            C_1(s, t, m) & = \begin{cases}\min _{s^{\prime}=s+1 \ldots t} C_{ck}\left(s, s^{\prime}, t, m\right) & m \geq m_{\varnothing}^{s, t} \\
            \infty & m<m_{\varnothing}^{s, t}\end{cases} \\
            C_2(s, t, m) & =\begin{cases}
            C_{all}(s, t, m) & m \geq m_{all}^{s, t} \\
            \infty & m<m_{all}^{s, t}
            \end{cases}
        \end{aligned}
    \end{equation}

    the terms $C_{ck}\left(s, s^{\prime}, t, m\right), C_{all}(s, t, m)$ and $m_{all}^{s, t}, m_{\varnothing}^{s, t}$ mentioned above are given by

    \begin{equation}
        \begin{aligned}
            C_{ck}\left(s, s^{\prime}, t, m\right) & =\sum_{k=s}^{s^{\prime}-1} u_f^k+C_{B P}\left(s^{\prime}, t, m-\omega_a^{s^{\prime}-1}\right) \\
            & + C_{B P}\left(s, s^{\prime}-1, m\right) \\
            C_{all}(s, t, m) & =u_f^s + u_{fcomm}^s + C_{B P}\left(s+1, t, m-\omega_{\bar{a}}^s\right) \\
            & + u_b^s + u_{bcomm}^s
        \end{aligned}
    \end{equation}

    and

    \begin{equation}
        \begin{aligned}
            & m_{\varnothing}^{s, t}=\max \begin{cases}
            \omega_\delta^t+\omega_a^s+o_f^s + o_{fcomm}^s \\
            \omega_\delta^t+\max _{s+1 \leq j<t}\left\{\omega_a^{j-1}+\omega_a^j+o_f^j + o_{fcommm}^j\right\}\end{cases}\\
            & m_{a l l}^{s, t}=\max\begin{cases}
            \omega_\delta^t+\omega_{\bar{a}}^s+o_f^s + o_{fcomm}^s\\
            \omega_\delta^s+\omega_{\bar{a}}^s+o_b^s + o_{bcomm}^s
        \end{cases} \\
        \end{aligned}
    \end{equation}
    \label{thm:act_main_thm}
\end{theorem}

Following the same algorithm mentioned in Rotor\cite{herrmann2019optimal} and replacing the original formula as the one mentioned above, we could include overheads that happen during the communication operation after applying the intra-op parallel solver.

\subsubsection{Linearization}\label{linearization}

\begin{figure}[h!]
  \includegraphics[width=0.45\textwidth]{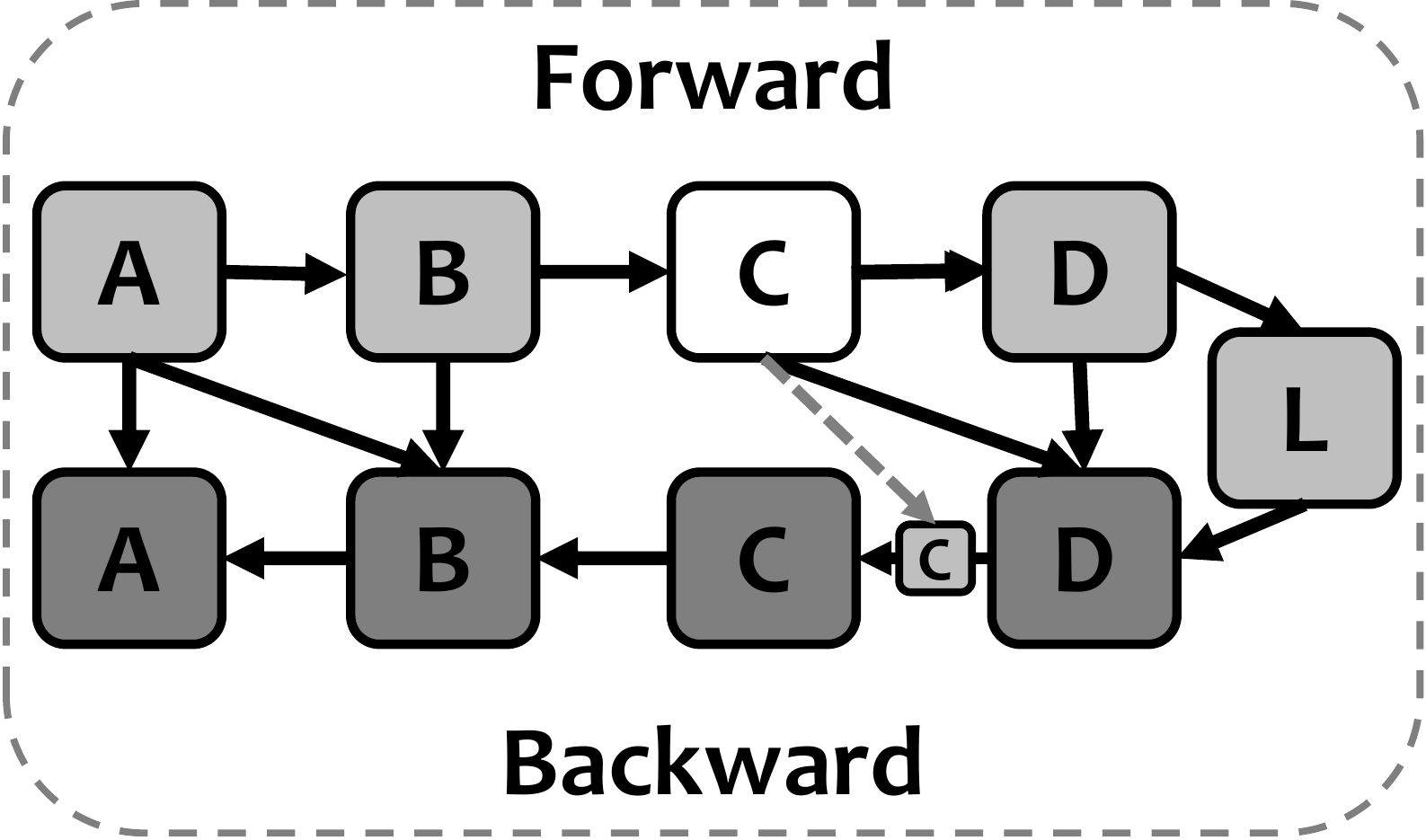}
  \caption{Linearized Computational Graph}
  \label{fig: arch}
\end{figure}

In the previous works \cite{herrmann2019optimal, beaumont2021efficient}, the network should follow the linearized assumption. This means that the network should have a sequential architecture where the output of a layer is only given to the next layer and not any other layer as shown in Figure~\ref{fig: arch}.

However, modern Deep Learning models usually don't naively follow the assumption. 
For example, there exist many residual connections in models such as ResNet~\cite{he2016deep} and Transformer~\cite{transformer}. 
To address the problem, previous works manually rewrite all the PyTorch models needed in their experiments to \texttt{torch.nn.Sequential}. 
However, it is laborsome for researchers and engineers to rewrite models to fit in this framework. 
Fortunately, with \texttt{torch.fx}, we are able to retrieve the forward computational graph and get access to node parents and children. 
Therefore, it is possible for us to do some dependency analysis on the graph structure to determine where to partition the graph to achieve this assumption. 
To clarify our basic rule for network linearization, we should define some of the components in this part.

\begin{definition}
\textit{We define some of the concepts related to network linearization here}
\begin{itemize}
    \item \textbf{Node}: \textit{node in the computational graph, represents an operation.}
    \item \textbf{Graph}:\textit{ computational graph of neural networks, and we assume it is a Directed Acyclic Graph (DAG).}
    \item \textbf{Node group}: \textit{the actual "node" for our solver, the neural network should be linearized with respect to node groups.}
    \item \textbf{End of a node group}: \textit{during the network linearization, we will examine nodes in the graph following the topological order, then append them to the current node group we are working on. The end of a node group means after appending this node into the working node group, we will move on to the next node group, i.e. the current node could be a partition point for the graph to achieve linearized assumption.
    \item \textbf{Existing dependencies}: a dynamic dependencies pool. While examing a node in the graph, we will first remove dependencies of this node's parents, and add dependencies of this node's children into this pool. The dependency is expressed by the number of the node's children, and removing dependency is expressed by subtracting the corresponding node's dependency by 1.}
\end{itemize}
\end{definition}
With the above definitions, our basic rule is simple: the node can remove all the existing dependencies, i.e. after removing its parents' dependencies, the existing dependencies are all 0, then this node is called the end of a node group. By doing so, it is possible to linearize classic residual networks such as ResNet-152~\cite{he2016deep}.

\subsubsection{Common Node}\label{common_node}

With the utilization of the aforementioned techniques, our solver is able to generate a linearized network automatically. However, the recent transformer-based networks pose a challenge in terms of linearization using the aforementioned basic rule. These networks typically involve constants such as attention masks within their training pipeline, which are used extensively by numerous nodes in the computational graph. Therefore, it is not feasible to linearize such models effectively without specifically processing these nodes. In fact, applying the basic rule to transformer-based networks would result in a node group containing all nodes.

A method for handling this issue was introduced by \cite{lai2022merak}, which involves the use of a common node mechanism to eliminate dependencies associated with nodes like attention masks, thereby enabling linearization of the network. However, their approach to identifying common nodes involves a simplistic threshold-based classification of nodes having a greater number of children than the given threshold. In order to design a more effective common node mechanism, it is necessary to clearly define the characteristics of nodes that can be considered common nodes. The definition of a common node is given below.

\begin{definition}
    \textit{We labeled a node as a common node if it is non-differentiable. For example, the attention mask in PyTorch is non-differentiable as the attention mask is \texttt{bool} type, so the gradient of this kind of node will not be created during training; or operations like \texttt{getattr} and \texttt{getitem} in Python are also non-differentiable.}
    \label{def: common-node}
\end{definition}

As we define the common node, we could further define the propagation of this kind of property throughout the network

\begin{lemma}
    Any node in the graph will be labeled as a common node if any of the following conditions hold:

    \begin{itemize}
        \item All of the parent nodes of this node are labeled as a common node
        \item The operation related to the node is non-differentiable, e.g. \texttt{getattr}, \texttt{getitem}, etc.
    \end{itemize}
    \label{lem:cnodeprop}
\end{lemma}

To initiate the common node propagation, we just need to label the original common node in network input, for example, the attention mask in transformer-based DNNs. Then, with the common node propagation, we are able to expand the common node set and remove dependencies of those nodes during network linearization.

\subsubsection{Full Set of Linearization Rules}

Considering the mechanism mentioned in \ref{linearization} and \ref{common_node}, we are about to give the full set of linearization rules. Before that, there are two more things that should be considered in the linearization process

\begin{itemize}
    \item As mentioned in \ref{modeling}, we include overheads related to communication in our modeling, but we don't need to explicitly add those overheads following the original formula, all we need is to stick those communication nodes to their parents, i.e. include communication nodes in the same group as their parents.
   \item The in-place operation is quite prevalent in modern DNNs, such as standard ResNet structure will set ReLU operation after Batch Normalization to in-place to save memory and time. Those operations together with their parents (those who don't need their output for the backward phase) could be combined so that the memory estimation will be clearer.
\end{itemize}

Therefore, here comes the final network linearization algorithm




\begin{algorithm}
\caption{network linearization algorithm}
\KwData{graph $\mathcal{G}$, common node-set $\mathcal{C}$}
\KwResult{linearized graph $\bar{\mathcal{G}}$}
\newcommand{\forcond}{$i=0$ \KwTo $n$}
\SetKwFunction{sink}{is\_sink}
\SetKwProg{Fn}{Function}{:}{}
\Fn{\sink{$deps\_pool, Node$}}{
\eIf{$deps\_pool$ is empty and all of $Node.children$ are not in-place operation or communication operation}{
\Return True\;
}{\Return False\;
}
}
$\bar{\mathcal{G}} \gets []$\;
$Current\_Node \gets []$\;
$deps\_pool \gets \{\}$\;
\For{$Node$ in $\mathcal{G}$}{
\If{$Node$ is not placeholder (i.e. input and output)}{
        \For{$Parent$ in $Node.parents$}{
            \If{$Parent$ is not placeholder and common node}{
                $deps\_pool[Parent] -= 1$\;
            }
        }
        append $Node$ to $Current\_Node$\;

        \If{\sink{$deps\_pool, Node$}}{
            append $Current\_Node$ to $\bar{\mathcal{G}}$\;
            $Current\_Node \gets []$\;
        }    
        \If{$Node$ satisfies \ref{lem:cnodeprop}}{
            append $Node$ to $\mathcal{C}$\;
        }{
            $deps\_pool[Node] \gets len(Node.children)$\;
        }
    }
}
\Return $\bar{\mathcal{G}}$\;
\end{algorithm}

\subsection{Integrate Two Solvers}

Due to the distinct approaches taken by the two solvers, their training processes' memory requirements are affected. Both solvers aim to optimize for the shortest possible execution time while remaining within the memory budget. If both solvers share the same memory budget, the intra-op parallelism solver's solution will compress the memory within the budget, rendering the activation checkpoint solver redundant. However, modeling the two problems together leads to a significant increase in complexity, making it infeasible to solve within an acceptable timeframe.

To address this issue, we utilize intra-op parallelism to solve the memory budget for a range of values $[(1+\alpha)^n]$ (where n is within the range [0, 9] and $\alpha$ denotes the expansion coefficient). In doing so, we solve a series of memory budgets and input the optimal solution under each budget into the activation checkpoint solver. Ultimately, our final execution plan is determined by identifying the execution plan with the shortest total execution time across different intra-op memory budgets.
\section{Generator}

The generator transforms the original non-distributed computation graph to one applied with the optimal execution plan given by our solver and it compiles the optimized computation graph to PyTorch code. To achieve this goal, we implement a set of compilation passes to manipulate the graph and a code generation feature to create PyTorch code based on optimized Graph IR.

\subsection{Compilation passes}

Our generator has a set of compilation passes to apply the searched execution plan.

Firstly, we design a \textit{communication-related pass} to insert communication nodes as required by the execution plan. Two kinds of communication nodes in our execution plan need to be considered: a) communication node to ensure numerical correctness, such as an all-reduce node after obtaining a partial sum output. b) communication node converting tensor to a layout required by a downstream user node.

Secondly, we design a \textit{parameter shard pass} to shard module parameters and add a hook to handle gradient communication. To improve training efficiency, we use an extra CUDA stream to overlap the asynchronous gradient communication with backward computation. \par

Finally, a \textit{reshape conversion pass} is used to adapt the argument of a node with reshape operators, such as transpose or reshape, to the searched execution plan. The reason we need this pass is that some nodes of the reshape operation use a constant that may come from the shape of the original input tensor, instead of the shard tensor in the execution plan.

\subsection{Code generation}

PyTorch FX~\cite{torch-fx} provides a code generation feature to generate valid Python code that adheres to the semantics of a given Graph. In order to support our automatic activation checkpoint feature, we made certain modifications to the PyTorch FX code generation.

Following the activation checkpoint searching process, we annotate each node with the activation checkpoint block index, based on the activation checkpoint solution. Nodes annotated with the same block index are then packaged together in an activation checkpoint block.

Our code generator initially generates a function that packages all the operations in the same activation checkpoint block. These generated functions are then wrapped in PyTorch's built-in checkpoint function, which enables input tensor preservation for backward computation and recomputation during forward computation. Finally, we substitute the nodes packaged in the activation checkpoint block in the original computation graph with the corresponding wrapped activation checkpoint function.
\section{Evaluation}

The benchmark is conducted with 8 Nvidia A100 (80GB) GPUs. It should be noted that only 4 pairs of adjacent GPUs are connected via NVLink, others are connected via PCI-e. 

\begin{table}[h]
    \centering
    \begin{tabular}{|c|c|c|c|}
        \hline
        experiment ID & \#GPUS & Hidden size & \#params(billion)\\
        \hline
        $\alpha$ & 1 & 2048 & 0.409\\
        \hline
        $\beta$ & 2 & 4096 & 1.221\\
        \hline
        $\gamma$ &4 & 8192 & 4.053\\
        \hline
        $\delta$ & 8 & 16384 & 14.550\\
        \hline
    \end{tabular}
    \caption{\textbf{GPT2 Model congiuration for experiments. We fix the number of layers as 4, and the sequence length as 1024. The reason we choose a relatively small layer number is that pipeline parallelism will split the model into partitions in which the number of transformer blocks is as small as our setting.}}
    \label{tab:experiment setting}
    \vspace{-1.9em}
\end{table}

For weak scaling, the problem size scales up as the number of devices increases. We compare our solution with Megatron-LM~\cite{shoeybi2019megatron} (1D tensor parallelism), Optimus~\cite{https://doi.org/10.48550/arxiv.2104.05343} (2D tensor parallelism) and ~\cite{https://doi.org/10.48550/arxiv.2105.14450} (3D tensor parallelism). The specific setting is shown in Tabel~\ref{tab:experiment setting}. The performance data is shown in Table~\ref{tab: experiment performance}. DDP is not included here because it encounters an out-of-memory issue when the problem size increases.

\begin{table}[h]
    \centering
    \begin{tabular}{|c|c|c|c|c|}
        \hline
        experiment ID & Megatron & Optimus & 3D TP & ours\\
        \hline
        $\alpha$ & 0.161 & 0.161 & 0.161 & 0.161\\
        \hline
        $\beta$ & 0.324 & - & - & 0.332\\
        \hline
        $\gamma$ & 0.528 & 0.368 & - & 0.604\\
        \hline
        $\delta$ & 0.728 & - & 0.715 & 0.824\\
        \hline
    \end{tabular}
    \caption{\textbf{Weak scaling performance. The model settings refer to Table~\ref{tab:experiment setting}. We measure the total PFLOPS for each training method.}}
    \label{tab: experiment performance}
    \vspace{-2em}
\end{table}




It can be observed that our system has the best performance in all cases, the reason is that our solver leverages the cluster information and the execution plan is adaptive to hardware configuration. For example, our execution plan for the experiment applies a data-parallelism-like strategy in devices across NUMA nodes and applies a tensor-parallelism-like strategy in 4 devices in a NUMA node. Our solution avoids the communication operations for tensor parallel via the cross NUMA PCI-e connection, and the communication operations via the lowest bandwidth connection could overlap with the backward computation.
\section{Future work}

\subsection{Construct strategy generator for operators}
In the intra-op parallelism stage, we need to generate all valid parallelism strategies for an operator. Although we use no more than 20 strategy generators to support the GPT2 model, it is still a big problem to support strategy generation for all 2000 operators in PyTorch. This problem will be eliminated after PyTorch2.0 releasing. As PrimTorch canonicalizes about 2000 PyTorch operators down to a closed set of about 250 primitive operators, it is possible for us to enumerate strategy generators for all operators.

\subsection{evaluation for integrating 2-stage solver}
The experiments to evaluate the performance of the 2-stage solver will be done in the future. We have already applied a 2-stage solver in Resnet50 model~\cite{he2016deep}. However, to better evaluate the performance on foundation models~\cite{bommasani2021opportunities}, there are still some trivial patching jobs.

\subsection{Support inter-op parallism}

Currently, we only search execution plans combining intra-op parallelism and activation checkpoint. We will add inter-op parallelism into our execution plan utilizing the same automatic methodology in the future. 
\section{Conclusion}


In this work, we introduce Colossal-Auto to jointly search for intra-op parallelism and activation checkpointing to generate the near-optimal distributed training plan for large-scale models. We use an analyzer to gather essential information about the hardware performance and the computation graph cost, employ a two-stage solver to find the near-optimal execution plan, and finally recompile the PyTorch model to a distributed model. 
Finding an efficient distributed training strategy is a challenging task that required extensive expert experience. 
We believe our system could lower the barrier for large model training for users without distributed training experience. 
Our future work could improve the robustness and versatility of this system further.


\newpage
\bibliographystyle{plainnat}
\bibliography{tex/sample-base}

\begin{thebibliography}{37}
\providecommand{\natexlab}[1]{#1}
\providecommand{\url}[1]{\texttt{#1}}
\expandafter\ifx\csname urlstyle\endcsname\relax
  \providecommand{\doi}[1]{doi: #1}\else
  \providecommand{\doi}{doi: \begingroup \urlstyle{rm}\Url}\fi

\bibitem[authors(2021)]{FairScale2021}
FairScale authors.
\newblock Fairscale: A general purpose modular pytorch library for high
  performance and large scale training.
\newblock \url{https://github.com/facebookresearch/fairscale}, 2021.

\bibitem[Beaumont et~al.(2021)Beaumont, Eyraud-Dubois, and
  Shilova]{beaumont2021efficient}
Olivier Beaumont, Lionel Eyraud-Dubois, and Alena Shilova.
\newblock Efficient combination of rematerialization and offloading for
  training dnns.
\newblock \emph{Advances in Neural Information Processing Systems},
  34:\penalty0 23844--23857, 2021.

\bibitem[Bian et~al.(2021)Bian, Xu, Wang, and
  You]{https://doi.org/10.48550/arxiv.2105.14450}
Zhengda Bian, Qifan Xu, Boxiang Wang, and Yang You.
\newblock Maximizing parallelism in distributed training for huge neural
  networks, 2021.

\bibitem[Bommasani et~al.(2021)Bommasani, Hudson, Adeli, Altman, Arora, von
  Arx, Bernstein, Bohg, Bosselut, Brunskill,
  et~al.]{bommasani2021opportunities}
Rishi Bommasani, Drew~A Hudson, Ehsan Adeli, Russ Altman, Simran Arora, Sydney
  von Arx, Michael~S Bernstein, Jeannette Bohg, Antoine Bosselut, Emma
  Brunskill, et~al.
\newblock On the opportunities and risks of foundation models.
\newblock \emph{arXiv preprint arXiv:2108.07258}, 2021.

\bibitem[Brown et~al.(2020)Brown, Mann, Ryder, Subbiah, Kaplan, Dhariwal,
  Neelakantan, Shyam, Sastry, Askell, Agarwal, Herbert-Voss, Krueger, Henighan,
  Child, Ramesh, Ziegler, Wu, Winter, Hesse, Chen, Sigler, Litwin, Gray, Chess,
  Clark, Berner, McCandlish, Radford, Sutskever, and Amodei]{gpt-3}
Tom~B. Brown, Benjamin Mann, Nick Ryder, Melanie Subbiah, Jared Kaplan,
  Prafulla Dhariwal, Arvind Neelakantan, Pranav Shyam, Girish Sastry, Amanda
  Askell, Sandhini Agarwal, Ariel Herbert-Voss, Gretchen Krueger, Tom Henighan,
  Rewon Child, Aditya Ramesh, Daniel~M. Ziegler, Jeffrey Wu, Clemens Winter,
  Christopher Hesse, Mark Chen, Eric Sigler, Mateusz Litwin, Scott Gray,
  Benjamin Chess, Jack Clark, Christopher Berner, Sam McCandlish, Alec Radford,
  Ilya Sutskever, and Dario Amodei.
\newblock Language models are few-shot learners, 2020.
\newblock URL \url{https://arxiv.org/abs/2005.14165}.

\bibitem[Chen et~al.(2016)Chen, Xu, Zhang, and Guestrin]{act-ckpt}
Tianqi Chen, Bing Xu, Chiyuan Zhang, and Carlos Guestrin.
\newblock Training deep nets with sublinear memory cost.
\newblock \emph{arXiv preprint arXiv:1604.06174}, 2016.

\bibitem[Devlin et~al.(2019)Devlin, Chang, Lee, and Toutanova]{bert}
Jacob Devlin, Ming-Wei Chang, Kenton Lee, and Kristina Toutanova.
\newblock {BERT}: Pre-training of deep bidirectional transformers for language
  understanding.
\newblock In \emph{Proceedings of the 2019 Conference of the North {A}merican
  Chapter of the Association for Computational Linguistics: Human Language
  Technologies, Volume 1 (Long and Short Papers)}, pages 4171--4186,
  Minneapolis, Minnesota, June 2019. Association for Computational Linguistics.
\newblock \doi{10.18653/v1/N19-1423}.
\newblock URL \url{https://aclanthology.org/N19-1423}.

\bibitem[Dosovitskiy et~al.(2020)Dosovitskiy, Beyer, Kolesnikov, Weissenborn,
  Zhai, Unterthiner, Dehghani, Minderer, Heigold, Gelly, Uszkoreit, and
  Houlsby]{https://doi.org/10.48550/arxiv.2010.11929}
Alexey Dosovitskiy, Lucas Beyer, Alexander Kolesnikov, Dirk Weissenborn,
  Xiaohua Zhai, Thomas Unterthiner, Mostafa Dehghani, Matthias Minderer, Georg
  Heigold, Sylvain Gelly, Jakob Uszkoreit, and Neil Houlsby.
\newblock An image is worth 16x16 words: Transformers for image recognition at
  scale, 2020.
\newblock URL \url{https://arxiv.org/abs/2010.11929}.

\bibitem[Fang et~al.(2022)Fang, Zhu, Li, Su, Yu, Zhou, and
  You]{fang2022parallel}
Jiarui Fang, Zilin Zhu, Shenggui Li, Hui Su, Yang Yu, Jie Zhou, and Yang You.
\newblock Parallel training of pre-trained models via chunk-based dynamic
  memory management.
\newblock \emph{IEEE Transactions on Parallel and Distributed Systems},
  34\penalty0 (1):\penalty0 304--315, 2022.

\bibitem[He et~al.(2016)He, Zhang, Ren, and Sun]{he2016deep}
Kaiming He, Xiangyu Zhang, Shaoqing Ren, and Jian Sun.
\newblock Deep residual learning for image recognition.
\newblock In \emph{Proceedings of the IEEE conference on computer vision and
  pattern recognition}, pages 770--778, 2016.

\bibitem[Herrmann et~al.(2019)Herrmann, Beaumont, Eyraud-Dubois, Hermann, Joly,
  and Shilova]{herrmann2019optimal}
Julien Herrmann, Olivier Beaumont, Lionel Eyraud-Dubois, Julien Hermann, Alexis
  Joly, and Alena Shilova.
\newblock Optimal checkpointing for heterogeneous chains: how to train deep
  neural networks with limited memory.
\newblock \emph{arXiv preprint arXiv:1911.13214}, 2019.

\bibitem[Huang et~al.(2022)Huang, Fang, Liu, Li, and You]{huang2022elixir}
Haichen Huang, Jiarui Fang, Hongxin Liu, Shenggui Li, and Yang You.
\newblock Elixir: Train a large language model on a small gpu cluster.
\newblock \emph{arXiv preprint arXiv:2212.05339}, 2022.

\bibitem[Huang et~al.(2019)Huang, Cheng, Bapna, Firat, Chen, Chen, Lee, Ngiam,
  Le, Wu, and Chen]{gpipe}
Yanping Huang, Youlong Cheng, Ankur Bapna, Orhan Firat, Mia~Xu Chen, Dehao
  Chen, HyoukJoong Lee, Jiquan Ngiam, Quoc~V. Le, Yonghui Wu, and Zhifeng Chen.
\newblock \emph{GPipe: Efficient Training of Giant Neural Networks Using
  Pipeline Parallelism}.
\newblock Curran Associates Inc., Red Hook, NY, USA, 2019.

\bibitem[Jain et~al.(2020)Jain, Jain, Nrusimha, Gholami, Abbeel, Gonzalez,
  Keutzer, and Stoica]{jain2020checkmate}
Paras Jain, Ajay Jain, Aniruddha Nrusimha, Amir Gholami, Pieter Abbeel, Joseph
  Gonzalez, Kurt Keutzer, and Ion Stoica.
\newblock Checkmate: Breaking the memory wall with optimal tensor
  rematerialization.
\newblock \emph{Proceedings of Machine Learning and Systems}, 2:\penalty0
  497--511, 2020.

\bibitem[Jia et~al.(2019)Jia, Zaharia, and Aiken]{jia2019beyond}
Zhihao Jia, Matei Zaharia, and Alex Aiken.
\newblock Beyond data and model parallelism for deep neural networks.
\newblock \emph{Proceedings of Machine Learning and Systems}, 1:\penalty0
  1--13, 2019.

\bibitem[Kirisame et~al.(2020)Kirisame, Lyubomirsky, Haan, Brennan, He, Roesch,
  Chen, and Tatlock]{dtr}
Marisa Kirisame, Steven Lyubomirsky, Altan Haan, Jennifer Brennan, Mike He,
  Jared Roesch, Tianqi Chen, and Zachary Tatlock.
\newblock Dynamic tensor rematerialization.
\newblock \emph{CoRR}, abs/2006.09616, 2020.
\newblock URL \url{https://arxiv.org/abs/2006.09616}.

\bibitem[Lai et~al.(2022)Lai, Li, Tang, Ge, Liu, Duan, Qiao, and
  Li]{lai2022merak}
Zhiquan Lai, Shengwei Li, Xudong Tang, Keshi Ge, Weijie Liu, Yabo Duan, Linbo
  Qiao, and Dongsheng Li.
\newblock Merak: A efficient distributed dnn training framework with automated
  3d parallelism for giant foundation models.
\newblock \emph{arXiv preprint arXiv:2206.04959}, 2022.

\bibitem[Lepikhin et~al.(2021)Lepikhin, Lee, Xu, Chen, Firat, Huang, Krikun,
  Shazeer, and Chen]{gshard}
Dmitry Lepikhin, HyoukJoong Lee, Yuanzhong Xu, Dehao Chen, Orhan Firat, Yanping
  Huang, Maxim Krikun, Noam Shazeer, and Zhifeng Chen.
\newblock Gshard: Scaling giant models with conditional computation and
  automatic sharding.
\newblock In \emph{9th International Conference on Learning Representations,
  {ICLR} 2021, Virtual Event, Austria, May 3-7, 2021}. OpenReview.net, 2021.
\newblock URL \url{https://openreview.net/forum?id=qrwe7XHTmYb}.

\bibitem[Li et~al.(2020)Li, Zhao, Varma, Salpekar, Noordhuis, Li, Paszke,
  Smith, Vaughan, Damania, and Chintala]{pytorch-distributed}
Shen Li, Yanli Zhao, Rohan Varma, Omkar Salpekar, Pieter Noordhuis, Teng Li,
  Adam Paszke, Jeff Smith, Brian Vaughan, Pritam Damania, and Soumith Chintala.
\newblock Pytorch distributed: Experiences on accelerating data parallel
  training.
\newblock \emph{Proc. VLDB Endow.}, 13\penalty0 (12):\penalty0 3005–3018, sep
  2020.
\newblock ISSN 2150-8097.
\newblock \doi{10.14778/3415478.3415530}.
\newblock URL \url{https://doi.org/10.14778/3415478.3415530}.

\bibitem[Li and Hoefler(2021)]{chimera}
Shigang Li and Torsten Hoefler.
\newblock Chimera: Efficiently training large-scale neural networks with
  bidirectional pipelines.
\newblock In \emph{Proceedings of the International Conference for High
  Performance Computing, Networking, Storage and Analysis}, SC '21, New York,
  NY, USA, 2021. Association for Computing Machinery.
\newblock ISBN 9781450384421.
\newblock \doi{10.1145/3458817.3476145}.
\newblock URL \url{https://doi.org/10.1145/3458817.3476145}.

\bibitem[Li et~al.(2021)Li, Zhuang, Guo, Zhuo, Zhang, Song, and
  Stoica]{terapipe}
Zhuohan Li, Siyuan Zhuang, Shiyuan Guo, Danyang Zhuo, Hao Zhang, Dawn Song, and
  Ion Stoica.
\newblock Terapipe: Token-level pipeline parallelism for training large-scale
  language models, 2021.
\newblock URL \url{https://arxiv.org/abs/2102.07988}.

\bibitem[Lu et~al.(2017)Lu, Yan, Li, Gong, Han, and Li]{flexflow}
Wenyan Lu, Guihai Yan, Jiajun Li, Shijun Gong, Yinhe Han, and Xiaowei Li.
\newblock Flexflow: A flexible dataflow accelerator architecture for
  convolutional neural networks.
\newblock In \emph{2017 IEEE International Symposium on High Performance
  Computer Architecture (HPCA)}, pages 553--564, 2017.
\newblock \doi{10.1109/HPCA.2017.29}.

\bibitem[Narayanan et~al.(2019)Narayanan, Harlap, Phanishayee, Seshadri,
  Devanur, Ganger, Gibbons, and Zaharia]{pipedream}
Deepak Narayanan, Aaron Harlap, Amar Phanishayee, Vivek Seshadri, Nikhil~R.
  Devanur, Gregory~R. Ganger, Phillip~B. Gibbons, and Matei Zaharia.
\newblock Pipedream: Generalized pipeline parallelism for dnn training.
\newblock In \emph{Proceedings of the 27th ACM Symposium on Operating Systems
  Principles}, SOSP '19, page 1–15, New York, NY, USA, 2019. Association for
  Computing Machinery.
\newblock ISBN 9781450368735.
\newblock \doi{10.1145/3341301.3359646}.
\newblock URL \url{https://doi.org/10.1145/3341301.3359646}.

\bibitem[Paszke et~al.(2019)Paszke, Gross, Massa, Lerer, Bradbury, Chanan,
  Killeen, Lin, Gimelshein, Antiga, et~al.]{paszke2019pytorch}
Adam Paszke, Sam Gross, Francisco Massa, Adam Lerer, James Bradbury, Gregory
  Chanan, Trevor Killeen, Zeming Lin, Natalia Gimelshein, Luca Antiga, et~al.
\newblock Pytorch: An imperative style, high-performance deep learning library.
\newblock \emph{Advances in neural information processing systems}, 32, 2019.

\bibitem[Patil et~al.(2022)Patil, Jain, Dutta, Stoica, and
  Gonzalez]{patil2022poet}
Shishir~G Patil, Paras Jain, Prabal Dutta, Ion Stoica, and Joseph Gonzalez.
\newblock Poet: Training neural networks on tiny devices with integrated
  rematerialization and paging.
\newblock In \emph{International Conference on Machine Learning}, pages
  17573--17583. PMLR, 2022.

\bibitem[Rajbhandari et~al.(2020)Rajbhandari, Rasley, Ruwase, and
  He]{rajbhandari2020zero}
Samyam Rajbhandari, Jeff Rasley, Olatunji Ruwase, and Yuxiong He.
\newblock Zero: Memory optimizations toward training trillion parameter models.
\newblock In \emph{SC20: International Conference for High Performance
  Computing, Networking, Storage and Analysis}, pages 1--16. IEEE, 2020.

\bibitem[Rajbhandari et~al.(2021)Rajbhandari, Ruwase, Rasley, Smith, and
  He]{zero-infinity}
Samyam Rajbhandari, Olatunji Ruwase, Jeff Rasley, Shaden Smith, and Yuxiong He.
\newblock Zero-infinity: Breaking the gpu memory wall for extreme scale deep
  learning, 2021.
\newblock URL \url{https://arxiv.org/abs/2104.07857}.

\bibitem[Rasley et~al.(2020)Rasley, Rajbhandari, Ruwase, and He]{deepspeed}
Jeff Rasley, Samyam Rajbhandari, Olatunji Ruwase, and Yuxiong He.
\newblock Deepspeed: System optimizations enable training deep learning models
  with over 100 billion parameters.
\newblock In \emph{Proceedings of the 26th ACM SIGKDD International Conference
  on Knowledge Discovery \& Data Mining}, KDD '20, page 3505–3506, New York,
  NY, USA, 2020. Association for Computing Machinery.
\newblock ISBN 9781450379984.
\newblock \doi{10.1145/3394486.3406703}.
\newblock URL \url{https://doi.org/10.1145/3394486.3406703}.

\bibitem[Reed et~al.(2021)Reed, DeVito, He, Ussery, and Ansel]{torch-fx}
James~K. Reed, Zachary DeVito, Horace He, Ansley Ussery, and Jason Ansel.
\newblock Torch.fx: Practical program capture and transformation for deep
  learning in python, 2021.
\newblock URL \url{https://arxiv.org/abs/2112.08429}.

\bibitem[Ren et~al.(2021)Ren, Rajbhandari, Aminabadi, Ruwase, Yang, Zhang, Li,
  and He]{ren2021zero}
Jie Ren, Samyam Rajbhandari, Reza~Yazdani Aminabadi, Olatunji Ruwase, Shuangyan
  Yang, Minjia Zhang, Dong Li, and Yuxiong He.
\newblock Zero-offload: Democratizing billion-scale model training.
\newblock In \emph{USENIX Annual Technical Conference}, pages 551--564, 2021.

\bibitem[Sergeev and Balso(2018)]{horovod}
Alexander Sergeev and Mike~Del Balso.
\newblock Horovod: fast and easy distributed deep learning in {TensorFlow}.
\newblock \emph{arXiv preprint arXiv:1802.05799}, 2018.

\bibitem[Shoeybi et~al.(2019)Shoeybi, Patwary, Puri, LeGresley, Casper, and
  Catanzaro]{shoeybi2019megatron}
Mohammad Shoeybi, Mostofa Patwary, Raul Puri, Patrick LeGresley, Jared Casper,
  and Bryan Catanzaro.
\newblock Megatron-lm: Training multi-billion parameter language models using
  model parallelism.
\newblock \emph{arXiv preprint arXiv:1909.08053}, 2019.

\bibitem[Vaswani et~al.(2017)Vaswani, Shazeer, Parmar, Uszkoreit, Jones, Gomez,
  Kaiser, and Polosukhin]{transformer}
Ashish Vaswani, Noam Shazeer, Niki Parmar, Jakob Uszkoreit, Llion Jones,
  Aidan~N Gomez, \L~ukasz Kaiser, and Illia Polosukhin.
\newblock Attention is all you need.
\newblock In I.~Guyon, U.~Von Luxburg, S.~Bengio, H.~Wallach, R.~Fergus,
  S.~Vishwanathan, and R.~Garnett, editors, \emph{Advances in Neural
  Information Processing Systems}, volume~30. Curran Associates, Inc., 2017.
\newblock URL
  \url{https://proceedings.neurips.cc/paper/2017/file/3f5ee243547dee91fbd053c1c4a845aa-Paper.pdf}.

\bibitem[Wang et~al.(2019)Wang, Huang, and Li]{tofu}
Minjie Wang, Chien-chin Huang, and Jinyang Li.
\newblock Supporting very large models using automatic dataflow graph
  partitioning.
\newblock In \emph{Proceedings of the Fourteenth EuroSys Conference 2019},
  EuroSys '19, New York, NY, USA, 2019. Association for Computing Machinery.
\newblock ISBN 9781450362818.
\newblock \doi{10.1145/3302424.3303953}.
\newblock URL \url{https://doi.org/10.1145/3302424.3303953}.

\bibitem[Xu et~al.(2021{\natexlab{a}})Xu, Li, Gong, and
  You]{https://doi.org/10.48550/arxiv.2104.05343}
Qifan Xu, Shenggui Li, Chaoyu Gong, and Yang You.
\newblock An efficient 2d method for training super-large deep learning models,
  2021{\natexlab{a}}.

\bibitem[Xu et~al.(2021{\natexlab{b}})Xu, Lee, Chen, Hechtman, Huang, Joshi,
  Krikun, Lepikhin, Ly, Maggioni, et~al.]{xu2021gspmd}
Yuanzhong Xu, HyoukJoong Lee, Dehao Chen, Blake Hechtman, Yanping Huang, Rahul
  Joshi, Maxim Krikun, Dmitry Lepikhin, Andy Ly, Marcello Maggioni, et~al.
\newblock Gspmd: general and scalable parallelization for ml computation
  graphs.
\newblock \emph{arXiv preprint arXiv:2105.04663}, 2021{\natexlab{b}}.

\bibitem[Zheng et~al.(2022)Zheng, Li, Zhang, Zhuang, Chen, Huang, Wang, Xu,
  Zhuo, Xing, et~al.]{zheng2022alpa}
Lianmin Zheng, Zhuohan Li, Hao Zhang, Yonghao Zhuang, Zhifeng Chen, Yanping
  Huang, Yida Wang, Yuanzhong Xu, Danyang Zhuo, Eric~P Xing, et~al.
\newblock Alpa: Automating inter-and $\{$Intra-Operator$\}$ parallelism for
  distributed deep learning.
\newblock In \emph{16th USENIX Symposium on Operating Systems Design and
  Implementation (OSDI 22)}, pages 559--578, 2022.

\end{thebibliography}


\end{document}